\documentclass{article}

\usepackage[final]{nips_arxiv}

\usepackage[utf8]{inputenc} 
\usepackage[T1]{fontenc}    
\usepackage{hyperref}       
\usepackage{subcaption}
\usepackage{url}            
\usepackage{booktabs}       
\usepackage{amsfonts}       
\usepackage{nicefrac}       
\usepackage{microtype}      
\usepackage{amsmath,amsfonts,amssymb,amsthm}
\usepackage{xparse}
\DeclareDocumentCommand{\citetm}{ O{} O{} m }{\citet[#1][#2]{#3}}
\DeclareDocumentCommand{\citeta}{ O{} O{} m }{\citet[#1][#2]{#3}}
\DeclareDocumentCommand{\citepm}{ O{} O{} m }{\citep[#1][#2]{#3}}
\DeclareDocumentCommand{\citepa}{ O{} O{} m }{\citep[#1][#2]{#3}}

\usepackage{graphicx}
\usepackage{ifthen}
\usepackage[utf8]{inputenc}
\usepackage{color}

\newcommand\sX{\ensuremath{\mathcal{X}}}
\newcommand\sY{\ensuremath{\mathcal{Y}}}

\newcommand\bE{\ensuremath{\mathbf{E}}}

\newcommand\bP{\ensuremath{\mathbf{P}}}

\newcommand\bR{\ensuremath{\mathbf{R}}}

\DeclareMathOperator*{\diag}{diag} 
\newcommand\p[1]{\ensuremath{\left( #1 \right)}} 

\newcommand\eqdef{\ensuremath{\stackrel{\rm def}{=}}} 

\ifthenelse{\isundefined{\definition}}{\newtheorem{definition}{Definition}}{}
\ifthenelse{\isundefined{\assumption}}{\newtheorem{assumption}{Assumption}}{}
\ifthenelse{\isundefined{\hypothesis}}{\newtheorem{hypothesis}{Hypothesis}}{}
\ifthenelse{\isundefined{\proposition}}{\newtheorem{proposition}{Proposition}}{}
\ifthenelse{\isundefined{\theorem}}{\newtheorem{theorem}{Theorem}}{}
\ifthenelse{\isundefined{\lemma}}{\newtheorem{lemma}{Lemma}}{}
\ifthenelse{\isundefined{\corollary}}{\newtheorem{corollary}{Corollary}}{}
\ifthenelse{\isundefined{\alg}}{\newtheorem{alg}{Algorithm}}{}
{\theoremstyle{definition}
  \ifthenelse{\isundefined{\example}}{\newtheorem{example}{Example}}{}
  \ifthenelse{\isundefined{\extension}}{\newtheorem{extension}{Extension}}{}
}

\def\[#1\]{\begin{align}#1\end{align}}
\def\(#1\){\begin{align*}#1\end{align*}}

\def\argmin{\operatornamewithlimits{arg\,min}}

\newcommand{\bprf}{\begin{proof}}
\newcommand{\eprf}{\end{proof}}
\newcommand{\blem}{\begin{lemma}}
\newcommand{\elem}{\end{lemma}}

\newcommand{\oo}{\mathcal{O}}

\usepackage{natbib}
\definecolor{mydarkblue}{rgb}{0,0.08,0.45}                                         
\hypersetup{
  colorlinks=true,
  linkcolor=mydarkblue,
  citecolor=mydarkblue,
  filecolor=mydarkblue,
  urlcolor=mydarkblue,
}
\setcitestyle{authoryear,round,citesep={;},aysep={,},yysep={;}}
\usepackage{tikz}
\usetikzlibrary{patterns,positioning,fit,calc}
\usepackage{paralist}
\usepackage{enumitem}
\usepackage{mathtools}
\usepackage{algorithm,algorithmicx,algcompatible,algpseudocode}
\usepackage{bbm}

\newcommand{\fv}{h}

\DeclareMathOperator{\Sym}{Sym}
\DeclareMathOperator{\poly}{poly}
\DeclareMathOperator{\gap}{gap}

\newcommand{\hE}{\hat{\bE}}
\newcommand{\thePoly}{\poly\p{k,\pi_{\min}^{-1},\lambda^{-1},\tau}}

\newcommand{\risk}{R}
\newcommand{\riski}{\tilde{R}}
\newcommand{\loss}{L}

\renewcommand{\paragraph}[1]{\textbf{#1}}

\newcommand{\theTitle}{Unsupervised Risk Estimation Using Only Conditional Independence Structure}
\title{\theTitle}

\author{
  Jacob Steinhardt \\
  Computer Science Department \\
  Stanford University \\
  {\tt jsteinhardt@cs.stanford.edu}
\And
	Percy Liang \\
  Computer Science Department \\
  Stanford University \\
  {\tt pliang@cs.stanford.edu}
}

\begin{document}

\maketitle

\begin{abstract}
We show how to estimate a model's test error from unlabeled data, on 
distributions very different from the training distribution, while 
assuming only that certain conditional independencies are preserved 
between train and test. We do not need to assume 
that the optimal predictor is the same between train and test,
or that the true distribution lies in any parametric family. We can 
also efficiently differentiate the error estimate to perform unsupervised discriminative learning. 
Our technical tool is the method of moments, which allows 
us to exploit conditional independencies in the absence of 
a fully-specified model.
Our framework encompasses a large family of losses including the 
log and exponential loss, and extends to structured output 
settings such as hidden Markov models.


\end{abstract}

\section{Introduction}
\label{sec:introduction}

How can we assess the accuracy of a model when 
the test distribution is very different than the 
training distribution?
To address this question, we study the problem of \emph{unsupervised risk estimation} 
\citepm{donmez2010unsupervised}---that is, given a loss function $\loss(\theta; x,y)$ and a fixed model $\theta$,
estimate the risk
$\risk(\theta) \eqdef \bE_{x,y \sim p^*}[\loss(\theta; x,y)]$ with respect to a
test distribution $p^*(x,y)$, given
access only to $m$ unlabeled examples $x^{(1:m)} \sim p^*(x)$. 
Unsupervised risk estimation lets us estimate model accuracy 
on a novel input distribution, and is thus important for building reliable 
machine learning systems.
Beyond evaluating a single model, it also provides a 
way of harnessing unlabeled data for learning: by minimizing the 
estimated risk over 
$\theta$, we can perform unsupervised 
learning and domain adaptation.

Unsupervised risk estimation is impossible without some 
assumptions on $p^*$, as otherwise $p^*(y \mid x)$---about which 
we have no observable information---could be arbitrary. 
How satisfied we should be with an estimator depends on 
how strong its underlying assumptions are. In this paper, we present 
an approach which rests on surprisingly weak assumptions---that
$p^*$ satisfies certain conditional 
independencies, but not that it lies in any parametric family or is 
close to the training distribution.

To give a flavor for our results, suppose that $y \in \{1,\ldots,k\}$ 
and that the loss decomposes as a sum of three parts: 
$L(\theta; x,y) = \sum_{v=1}^3 f_v(\theta; x_v, y)$, where $x_1$, $x_2$, and 
$x_3$ are independent conditioned on $y$. In this case, we can estimate the 
risk to error $\epsilon$ in $\poly(k)/\epsilon^2$ samples, independently 
of the dimension of $x$ or $\theta$; the dependence on $k$ is roughly cubic in 
practice. In Sections~\ref{sec:framework} and \ref{sec:extensions} we generalize 
this result to capture both the log and exponential losses, and extend beyond 
the multiclass case to allow $y$ to be the hidden state of a hidden Markov model.

{
\newcommand{\dbar}[3]{
  \draw[fill=red] ({#3*\W+#1*\x-0.5*\w},{\b}) rectangle ({#3*\W+#1*\x+0.5*\w},{\b+#2*\y});
  \node at ({#3*\W+#1*\x},{-\c}) {\small $#1$};
}
\newcommand{\dline}[1]{
  \draw ({#1*\W+1*\x-0.9*\w},0) -- ({#1*\W+4*\x+\w},0);
  \draw ({#1*\W+1*\x-0.9*\w},0) -- ({#1*\W+1*\x-0.9*\w},{2*\b+\y});
}
\newcommand{\dlabeli}[3]{
  \node at ({#2*\W+#1*\x},{-2.5*\c}) {#3};
}
\newcommand{\dlabelii}[2]{
  \node at ({#1*\W-0.25*\x},{\b+0.5*\y}) {#2};
}
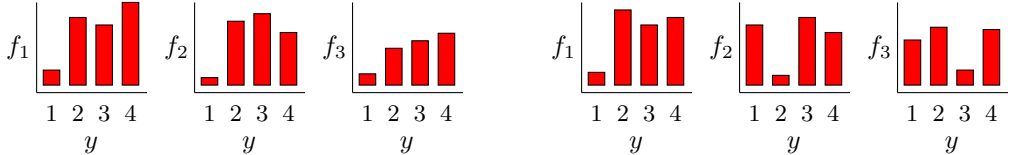
\begin{figure}
\begin{center}
\begin{tikzpicture}
\def\W{2.1}
\def\w{0.22}
\def\b{0.10}
\def\c{0.28}
\def\x{0.35}
\def\y{1}
\dbar{1}{0.2}{0}
\dbar{2}{0.9}{0}
\dbar{3}{0.8}{0}
\dbar{4}{1.1}{0}
\dbar{1}{0.1}{1}
\dbar{2}{0.85}{1}
\dbar{3}{0.95}{1}
\dbar{4}{0.7}{1}
\dbar{1}{0.15}{2}
\dbar{2}{0.49}{2}
\dbar{3}{0.59}{2}
\dbar{4}{0.69}{2}
\dline{0}
\dline{1}
\dline{2}
\dlabeli{2.5}{0}{$y$}
\dlabeli{2.5}{1}{$y$}
\dlabeli{2.5}{2}{$y$}
\dlabelii{0}{$f_1$}
\dlabelii{1}{$f_2$}
\dlabelii{2}{$f_3$}

\def\shift{0.45}
\dbar{1}{0.17}{(3+\shift)}
\dbar{2}{1.0} {(3+\shift)}
\dbar{3}{0.8} {(3+\shift)}
\dbar{4}{0.9} {(3+\shift)}

\dbar{1}{0.8} {(4+\shift)}
\dbar{2}{0.13}{(4+\shift)}
\dbar{3}{0.9} {(4+\shift)}
\dbar{4}{0.7} {(4+\shift)}

\dbar{1}{0.6} {(5+\shift)}
\dbar{2}{0.77}{(5+\shift)}
\dbar{3}{0.2} {(5+\shift)}
\dbar{4}{0.74}{(5+\shift)}
\dline{(3+\shift)}
\dline{(4+\shift)}
\dline{(5+\shift)}
\dlabeli{2.5}{(3+\shift)}{$y$}
\dlabeli{2.5}{(4+\shift)}{$y$}
\dlabeli{2.5}{(5+\shift)}{$y$}
\dlabelii{(3+\shift)}{$f_1$}
\dlabelii{(4+\shift)}{$f_2$}
\dlabelii{(5+\shift)}{$f_3$}
\dlabelii{(5.85+\shift)}{}

\end{tikzpicture}
\end{center}
\vskip -0.08in
\caption{Two possible loss profiles at a given 
value of $x$. Left: if $f_1$, $f_2$, and $f_3$ are all 
minimized at the same value of $y$, that is likely 
to be the correct value and the total loss is likely 
to be small. Right: conversely, if $f_1$, $f_2$, and $f_3$ 
are small at differing values of $y$, then the loss 
is likely to be large.}
\label{fig:bar-intuition}
\vskip -0.3in
\end{figure}
}

Some intuition is provided in Figure~\ref{fig:bar-intuition}. At a fixed value of 
$x$, we can think of each $f_v$ as ``predicting'' that $y=j$ if $f_v(x_v,j)$ is low and 
$f_v(x_v,j')$ is high for $j' \neq j$.
Since $f_1$, $f_2$, and $f_3$ all provide independent signals about $y$, their rate of agreement 
gives information about the model accuracy. If 
$f_1$, $f_2$, and $f_3$ all predict that $y = 1$,
then it is likely that the true $y$ equals $1$ and the 
loss is small. Conversely, if $f_1$, $f_2$, and $f_3$ all predict 
different values of $y$, then the loss is likely larger.
This intuition is formalized by \citetm{dawid1979maximum}
when the $f_v$ take values in a discrete set 
(e.g. when the $f_v$ measure the $0/1$-loss of independent classifiers); they model $(f_1,f_2,f_3)$ as a 
mixture of independent categorical variables, and use this to impute $y$ as the 
label of the mixture component.
Several others have extended this idea 
\citepm[e.g.][]{zhang2014crowdsourcing,platanios2015estimating,
jaffe2015estimating}, but continue to focus on the 
$0/1$ loss (with a single exception that we discuss below).

Why have continuous losses such as the log loss been ignored, given their 
utility for gradient-based learning? The issue is that 
while the $0/1$-loss only involves a discrete prediction in 
$\{1,\ldots,k\}$ (the predicted output), the log loss involves
predictions in $\bR^k$ (the predicted probability distribution over outputs). 
The former can be fully modeled by a $k$-dimensional family, while 
the latter requires infinitely many parameters. 
We could assume that the predictions are distributed according to some 
parametric family, but if that assumption is wrong then our risk estimates will 
be wrong as well.

To sidestep this issue, we make use of the \emph{method of moments}; while 
the method of moments has seen recent use in machine learning for fitting 
non-convex latent variable models \citepm[e.g.][]{anandkumar12moments}, it has a 
much older history in the econometrics literature, where it has been used as a 
tool for making causal identifications under structural assumptions, even when an 
explicit form for the likelihood is not known \citepm{anderson1949estimation,
anderson1950asymptotic,sargan1958estimation,sargan1959estimation, 
hansen1982gmm,powell1994estimation,hansen2014uncertainty}. 
It is upon this older literature 
that we draw conceptual inspiration, though our technical tools are more closely 
based on the newer machine learning approaches.
The key insight is that certain moment equations--e.g., $\bE[f_1f_2\mid y] = \bE[f_1\mid y]\bE[f_2 \mid y]$--
can be derived from the assumed independencies; we then show how to estimate the 
risk while relying only on these moment conditions, and not on any parametric 
assumptions about the $x_v$ or $f_v$. Moreover, these moment equations also 
hold for the gradient of $f_v$, which enables efficient unsupervised learning.

Our paper is structured as follows. In Section~\ref{sec:framework}, we 
present our basic framework, and state and prove our main result on estimating 
the risk given $f_1$, $f_2$, and $f_3$. In Section~\ref{sec:extensions}, we 
extend our framework in several directions, including to hidden 
Markov models. In Section~\ref{sec:learning}, we present a gradient-based 
learning algorithm and show that the sample complexity needed for learning 
is $d \cdot \poly(k) / \epsilon^2$, where $d$ is the dimension of $\theta$.
In Section~\ref{sec:experiments}, we investigate how our method performs 
empirically.

\paragraph{Related Work.}
While the formal problem of unsupervised risk estimation was only posed 
recently by \citetm{donmez2010unsupervised}, several older ideas from 
domain adaptation and semi-supervised learning are also relevant. 
The \emph{covariate shift assumption} assumes access to labeled samples from a base distribution 
$p_0(x,y)$ for which $p^*(y \mid x) = p_0(y \mid x)$.
If $p^*(x)$ and $p_0(x)$ are close together, 
we can approximate $p^*$ by $p_0$ via sample re-weighting 
\citepa{shimodaira2000improving,quinonero2009dataset}. 
If $p^*$ and $p_0$ are not close, another approach is to assume a well-specified discriminative model 
family $\Theta$, such that $p_0(y \mid x) = p^*(y \mid x) = p_{\theta^*}(y \mid x)$ 
for some $\theta^* \in \Theta$; then we need 
only heed finite-sample error in the estimation of $\theta^*$ 
\citepa{blitzer2011domain,li2011knows}.
Both assumptions are somewhat stringent --- re-weighting only 
allows small perturbations, and mis-specified models are common in practice. 
Indeed, many authors report that mis-specification
can lead to severe issues in semi-supervised settings 
\citepa{merialdo94tagging,nigam1998learning,cozman2006risks,liang08errors,li2015towards}.

As mentioned above, our approach is closer in spirit to that of 
\citetm{dawid1979maximum} and its extensions. 
Similarly to \citetm{zhang2014crowdsourcing} and \citetm{jaffe2015estimating}, 
we use the method of moments for estimating latent-variable models 
However, those papers use it as a tool for 
parameter estimation in the face of non-convexity, rather than as 
a way to sidestep model mis-specification. The insight that moments 
are robust to model mis-specification lets us 
extend beyond the simple discrete settings they consider in order to 
handle more complex continuous and structured losses.
Another approach to handling continuous losses is given in 
the intriguing work of \citetm{balasubramanian2011unsupervised}, who show 
that the distribution of losses $\loss \mid y$ is often close to Gaussian in practice, and use 
this to estimate the risk. 
A key difference from all of this prior work is that
we are the first to perform gradient-based learning and the 
first to handle a structured loss (in our case, the log loss for hidden Markov models).

\section{Framework and Estimation Algorithm}
\label{sec:framework}

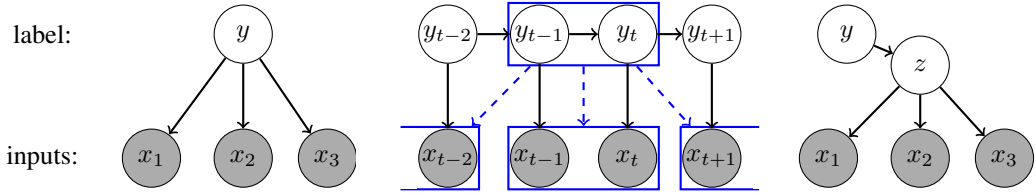
\begin{figure}
\begin{tikzpicture}
\def\A{4.55}
\def\a{-0.85}
\def\b{0.37}
\def\c{1.49}
\def\s{0}
\def\t{-1.65}
\node at (-2.3,\s) {label:};
\node at (-2.3,\t) {inputs:};

\begin{scope}[every node/.style={minimum size=2.2em,inner sep=0.2}]
\node[draw,circle] (y) at (\b,\s) {$y$};
\node[fill=gray!65,draw,circle] (x1) at (\a,\t) {$x_1$};
\node[fill=gray!65,draw,circle] (x2) at (\b,\t) {$x_2$};
\node[fill=gray!65,draw,circle] (x3) at (\c,\t) {$x_3$};
\draw[->,thick] (y) -- (x1);
\draw[->,thick] (y) -- (x2);
\draw[->,thick] (y) -- (x3);

\node[draw,circle] (y21) at (\A+1.5*\a-0.5*\b,\s) {$y_{t-2}$};
\node[draw,circle] (y22) at (\A+0.5*\a+0.5*\b,\s) {$y_{t-1}$};
\node[draw,circle] (y23) at (\A+0.5*\b+0.5*\c,\s) {$y_{t}$};
\node[draw,circle] (y24) at (\A+1.5*\c-0.5*\b,\s) {$y_{t+1}$};
\node[fill=gray!65,draw,circle] (x21) at (\A+1.5*\a-0.5*\b,\t) {$x_{t-2}$};
\node[fill=gray!65,draw,circle] (x22) at (\A+0.5*\a+0.5*\b,\t) {$x_{t-1}$};
\node[fill=gray!65,draw,circle] (x23) at (\A+0.5*\b+0.5*\c,\t) {$x_{t}$};
\node[fill=gray!65,draw,circle] (x24) at (\A+1.5*\c-0.5*\b,\t) {$x_{t+1}$};
\node (x25) at (\A+1.5*\c-0.5*\b+0.2,\t) {};
\node (x20) at (\A+1.5*\a-0.5*\b-0.2,\t) {};
\node (x2z) at (\A+1.5*\c-0.5*\b+0.95,\t) {};
\node (x2a) at (\A+1.5*\a-0.5*\b-0.81,\t) {};

\node[draw=blue,rectangle,thick,inner sep=0.6pt, fit=(y22) (y23)] (yp) {};
\node[draw=blue,rectangle,thick,inner sep=0.6pt, fit=(x21) (x20)] (x1p) {};
\node[draw=blue,rectangle,thick,inner sep=0.6pt, fit=(x22) (x23)] (x2p) {};
\node[draw=blue,rectangle,thick,inner sep=0.6pt, fit=(x24) (x25)] (x3p) {};
\node[fill=white,rectangle,inner sep=0.2pt, fit=(x2z)] {};
\node[fill=white,rectangle,inner sep=0.2pt, fit=(x2a)] {};

\draw[->,thick] (y21) -- (y22);
\draw[->,thick] (y22) -- (y23);
\draw[->,thick] (y23) -- (y24);
\draw[->,thick] (y21) -- (x21);
\draw[->,thick] (y22) -- (x22);
\draw[->,thick] (y23) -- (x23);
\draw[->,thick] (y24) -- (x24);
\draw[->,thick,dashed,color=blue] ($(yp.south west)!0.3!(yp.south)$) -- (x1p);
\draw[->,thick,dashed,color=blue] ($(yp.south east)!0.3!(yp.south)$) -- (x3p);
\draw[->,thick,dashed,color=blue] (yp) -- (x2p);

\node[draw,circle] (y3) at (-0.1+2*\A+0.8*\a+0.2*\b,\s) {$y$};
\node[draw,circle] (z3) at (-0.1+2*\A+\b,0.75*\s+0.25*\t) {$z$};
\node[fill=gray!65,draw,circle] (x31) at (-0.1+2*\A+\a,\t) {$x_1$};
\node[fill=gray!65,draw,circle] (x32) at (-0.1+2*\A+\b,\t) {$x_2$};
\node[fill=gray!65,draw,circle] (x33) at (-0.1+2*\A+\c,\t) {$x_3$};
\draw[->,thick] (y3) -- (z3);
\draw[->,thick] (z3) -- (x31);
\draw[->,thick] (z3) -- (x32);
\draw[->,thick] (z3) -- (x33);
\end{scope}

\end{tikzpicture}
\vskip -0.0in
\caption{Left: our basic $3$-view setup (Assumption~\ref{ass:multi-view}). 
Center: Extension~\ref{ex:hmm}, to hidden Markov models; 
the embedding of $3$ views into the HMM is indicated in blue. 
Right: Extension~\ref{ex:latent}, to include a mediating variable $z$.}
\label{fig:multiview}
\label{fig:hmm}
\label{fig:latent}
\vskip -0.15in
\end{figure}

We will focus on multiclass classification; we assume an unknown true distribution 
$p^*(x,y)$ over $\sX \times \sY$, where $\sY = \{1,\ldots,k\}$, and are given
unlabeled samples $x^{(1)}, \ldots, x^{(m)}$ drawn i.i.d.~from 
$p^*(x)$. Given parameters $\theta \in \mathbb R^d$ and a loss function $\loss(\theta; x,y)$, 
our goal is to estimate the risk of $\theta$ on $p^*$:
$\risk(\theta) \eqdef \bE_{x,y\sim p^*}[\loss(\theta; x,y)]$.
Throughout, we will make the \emph{3-view assumption}:

\begin{assumption}[3-view]
\label{ass:multi-view}
Under $p^*$, $x$ can be split into $x_1, x_2, x_3$, which are conditionally independent given $y$
(see Figure~\ref{fig:multiview}).
Moreover, the loss decomposes additively across views: 
$\loss(\theta; x,y) = A(\theta; x) - \sum_{v=1}^3 f_{v}(\theta; x_v,y)$,
for some functions $A$ and $f_v$.
\end{assumption}
Note that if we have $v > 3$ views $x_1,\ldots,x_v$,
then we can always partition the views into blocks 
$x'_1 = x_{1:\lfloor v/3 \rfloor}$, $x_2' = x_{\lfloor v/3\rfloor+1:\lfloor 2v/3 \rfloor}$, $x_3' = x_{\lfloor 2v/3 \rfloor+1:v}$. 
Assumption~\ref{ass:multi-view} then holds for $x_1', x_2', x_3'$.\footnote{
For $v = 2$ views, recovering $\risk$ is 
related to non-negative matrix factorization \citepa{lee2001algorithms}. 
Exact identification ends up being impossible, though 
obtaining upper bounds is likely possible.}
In addition, it suffices for just the $f_v$ to be independent rather than the $x_v$.

We will give some examples where Assumption~\ref{ass:multi-view} holds, 
then state and prove our main result. We start with logistic regression, 
which will be our primary focus later on:

\begin{example}[Logistic Regression]
\label{ex:logistic}
Suppose that we have a log-linear model 
$p_{\theta}(y \mid x) = \exp\p{\theta^{\top}\p{\phi_1(x_1,y) + \phi_2(x_2,y) + \phi_3(x_3,y)} - A(\theta;x)}$, where $x_1$, $x_2$, and $x_3$ 
are independent conditioned on $y$. 
If our loss function is the log-loss $\loss(\theta; x,y) = -\log p_{\theta}(y \mid x)$, then Assumption~\ref{ass:multi-view} holds with 
$f_v(\theta; x_v, y) = \theta^{\top}\phi_v(x_v,y)$, and $A(\theta; x)$ equal to 
the partition function of $p_{\theta}$.
\end{example}

We next consider the hinge loss, for which 
Assumption~\ref{ass:multi-view} does \emph{not} hold.
However, it does hold for a modified hinge loss, where we apply the hinge 
separately to each view:
\begin{example}[Modified Hinge Loss]
Suppose that $\loss(\theta; x,y) = \sum_{v=1}^3 (1 + \max_{j \neq y} \theta^{\top}\phi_v(x_v,j) - \theta^{\top}\phi_v(x_v,y))_+$. In other words, 
$\loss$ is the sum of $3$ hinge losses, one for each view. Then 
Assumption~\ref{ass:multi-view} holds with $A = 0$, and $-f_v$ equal to the 
hinge loss for view $v$.
\end{example}
There is nothing special about the hinge loss; for instance, we could 
instead take a sum of $0/1$-losses.

Our final example shows that linearity is not necessary 
for Assumption~\ref{ass:multi-view} to hold; the model can be 
arbitrarily non-linear in each view $x_v$, as long as the predictions are 
combined additively at the end:
\begin{example}[Neural Networks]
\label{ex:neural}
Suppose that for each view $v$ we have a neural network whose output 
is a prediction vector $(f_v(\theta; x_v, j))_{j=1}^k$. Suppose further 
that we add together the predictions $f_1+f_2+f_3$, apply a soft-max, and 
evaluate using the log loss; 
then $\loss(\theta; x,y) = A(\theta; x) - \sum_{v=1}^3 f_v(\theta; x_v, y)$, 
where $A(\theta; x)$ is the log-normalization constant of the softmax, and 
hence $\loss$ satisfies Assumption~\ref{ass:multi-view}.
\end{example}

With these examples in hand, we are ready for our main result on recovering 
the risk $\risk(\theta)$. 
The key is to recover the \emph{conditional risk matrices} $M_v \in \bR^{k \times k}$, defined as 
\begin{equation}
\label{eq:M-def}
(M_v)_{ij} = \bE[f_v(\theta; x_v, i) \mid y=j].
\end{equation}
%
In the case of the $0/1$-loss, the $M_v$ are
confusion matrices; in general, $(M_v)_{ij}$ measures how strongly we predict class $i$ when 
the true class is $j$.
If we could recover these matrices along with the marginal class 
probabilities $\pi_j \eqdef p^*(y=j)$, then estimating the risk would be straightforward; indeed, 
\vskip -0.17in
\begin{equation}
\label{eq:M-to-R}
\risk(\theta) = \bE[A(\theta; x)] - \sum_{j=1}^k \pi_{j} \sum_{v=1}^3 (M_v)_{j,j},
\end{equation}
\vskip -0.10in
where $\bE[A(\theta; x)]$ can be estimated from unlabeled data alone.

\paragraph{Caveat: Class permutation.} Suppose that at training time, 
we learn to predict whether an image contains the digit $0$ or $1$. At 
test time, nothing changes except the definitions of $0$ and $1$ are
reversed. It is clearly impossible to detect this from unlabeled data; 
mathematically, this manifests as $M_v$ only being recoverable up to 
column permutation. We will end up computing the minimum 
risk over these permutations, which we call the \emph{optimistic risk} 
and denote $\riski(\theta) \eqdef \min_{\sigma \in \Sym(k)} \bE_{x,y \sim p^*}[L(\theta; x, \sigma(y))]$.
This equals the true risk as long as $\theta$ is at least 
aligned with the correct labels in the sense that 
$\bE_x[L(\theta; x,j) \mid y=j] \leq \bE_x[L(\theta; x,j') \mid y=j]$ for $j' \neq j$.
The optimal $\sigma$ can then be computed from $M_v$ and $\pi$ in $\oo\p{k^3}$ time 
using maximum weight bipartite matching; 
see Section \ref{sec:matching-details} for details.

Our main result, Theorem~\ref{thm:tensor}, says that we can recover both 
$M_v$ and $\pi$ up to permutation, with a number of samples that is polynomial in $k$; in practice 
the dependence on $k$ seems roughly cubic.
\begin{theorem}
\label{thm:tensor}
Suppose Assumption~\ref{ass:multi-view} holds. 
Then, for any $\epsilon, \delta \in (0,1)$, we can estimate $M_v$ and $\pi$ up to column permutation,
to error $\epsilon$ (in Frobenius and $\infty$-norm respectively). Our algorithm requires
\begin{align}
\notag m =\, &\thePoly \cdot \frac{\log(2/\delta)}{\epsilon^2} \text{ samples to succeed with probability $1-\delta$, where} \\
\label{eq:estimation-complexity} 
\pi_{\min} &\eqdef \min_{j=1}^k p^*(y=j), \,\,\, \tau \eqdef \bE\big[{\textstyle\sum_{v,j}} f_v(\theta; x_v, j)^2\big], \text{ \ and \ } 
\lambda \eqdef \min_{v=1}^3 \sigma_{k}(M_v),
\end{align}
and $\sigma_k$ denotes the $k$th singular value.
Moreover, the algorithm runs in time $m \cdot \poly(k)$.
\end{theorem}
Note that estimates for $M_v$ and $\pi$ imply an estimate 
for $\riski$ via \eqref{eq:M-to-R}.
Importantly, the sample complexity in Theorem~\ref{thm:tensor} depends 
on the number of classes $k$, but not on the dimension $d$ of $\theta$. 
Moreover, Theorem~\ref{thm:tensor} holds even if 
$p^*$ lies outside the model family $\theta$, and even if the train and test 
distributions are very different (in fact, the result is totally agnostic 
to how the model $\theta$ was produced). The only requirement is that the 
$3$-view assumption holds for $p^*$.

Let us interpret each term in \eqref{eq:estimation-complexity}. First, $\tau$ 
tracks the variance of the loss, and we should expect the difficulty of 
estimating the risk to increase with this variance. 
The $\frac{\log(2/\delta)}{\epsilon^2}$ term is typical and shows up 
even when estimating the parameter of a Bernoulli variable 
to accuracy $\epsilon$ from $m$ samples. The $\pi_{\min}^{-1}$ term 
appears because, if one of the classes is very rare, we need to wait a 
long time to observe even a single sample from that class, and even longer 
to estimate the risk on that class accurately.

Perhaps least intuitive is 
the $\lambda^{-1}$ term, which is large e.g.~when two classes 
have similar conditional risk vectors $\bE[(f_v(\theta; x_v, i))_{i=1}^k \mid y=j]$.
To see why this matters, consider an extreme where $x_1$, $x_2$, and 
$x_3$ are independent not only of each other but also of $y$. Then 
$p^*(y)$ is completely unconstrained, and it is
impossible to estimate $\risk$ at all. Why 
does this not contradict Theorem~\ref{thm:tensor}? The answer is that in this 
case, all rows of $M_v$ are equal and hence $\lambda = 0$, 
$\lambda^{-1} = \infty$, and we need infinitely many samples for 
Theorem~\ref{thm:tensor} to hold; $\lambda$ thus 
measures how close we are to this degenerate case.

\paragraph{Proof of Theorem~\ref{thm:tensor}.}
We now outline a proof of Theorem~\ref{thm:tensor}. Recall 
the goal is to estimate the conditional risk matrices $M_v$, defined 
as $(M_v)_{ij} = \bE[f_v(\theta; x_v, i) \mid y=j]$; from these we can 
recover the risk itself using \eqref{eq:M-to-R}.
The key insight is that certain moments of $p^*(y \mid x)$ can be expressed 
as polynomial functions of the matrices $M_v$, and therefore we can solve 
for the $M_v$ even without explicitly estimating $p^*$.
Our approach follows the technical machinery behind the spectral method of moments 
\citepm[e.g.,][]{anandkumar12moments}, which 
we explain below for completeness.

Define the loss vector $\fv_v(x_v) = (f_v(\theta; x_v, i))_{i=1}^k$.
The conditional independence of the $x_v$ means that 
$\bE[\fv_1(x_1)\fv_2(x_2)^{\top} \mid y] = \bE[\fv_1(x_1) \mid y]\bE[\fv_2(x_2) \mid y]^{\top}$, and similarly for higher-order conditional moments. 
There is thus low-rank structure in the moments of $\fv$, which we can exploit. 
More precisely, by marginalizing over $y$, we obtain the 
following equations, where $\otimes$ denotes outer product:
\begin{align}
\notag \bE[&h_v(x_v)] = M_v\pi , \quad \bE[h_v(x_v)h_{v'}(x_{v'})^{\top}] = M_v\diag(\pi)M_{v'}^{\top} \text{ for $v \neq v'$, and } \\
\label{eq:moments} &\bE[h_1(x_1) \!\otimes\! h_2(x_2) \!\otimes\! h_3(x_3)]_{i_1,i_2,i_3} = \sum_{j=1}^k \pi_j \cdot (M_1)_{i_1,j} (M_2)_{i_2,j} (M_3)_{i_3,j}\,\, \forall i_1,i_2,i_3.
\end{align}
The left-hand-side of each equation can be estimated from 
unlabeled data; we can then solve for $M_v$ and $\pi$ using \emph{tensor 
decomposition} \citepa{delathauwer2006decomposition,comon2009tensor,
anandkumar12moments,anandkumar13tensor,kuleshov2015tensor}. 
In particular, we can recover $M$ and $\pi$ up to 
permutation: that is, we recover $\hat{M}$ and $\hat{\pi}$ such that
$M_{i,j} \approx \hat{M}_{i,\sigma(j)}$ and
$\pi_j \approx \hat{\pi}_{\sigma(j)}$
for some permutation $\sigma \in \Sym(k)$. 
This then yields Theorem~\ref{thm:tensor}; see Section~\ref{sec:tensor-proof} 
for a full proof.

Assumption~\ref{ass:multi-view} therefore yields a 
set of moment equations \eqref{eq:moments} that, when solved, let us estimate 
the risk without any labels $y$.
To summarize the procedure, we (i) approximate the left-hand-side of 
each term in \eqref{eq:moments} by sample averages; (ii) use tensor decomposition 
to solve for $\pi$ and $M_v$; (iii) use maximum matching to compute 
the permutation $\sigma$; and (iv) use \eqref{eq:M-to-R} to obtain 
$\riski$ from $\pi$ and $M_v$.

\setlength{\textfloatsep}{8pt}
\begin{algorithm}[t]
\caption{Algorithm for estimating $\riski(\theta)$ from unlabeled data.}
\label{alg:estimation}                                       
\begin{algorithmic}[1]                                        
\State \textbf{Input}: unlabeled samples $x^{(1)},\ldots,x^{(m)} \sim p^*(x)$.
\State Estimate the left-hand-side of each term in \eqref{eq:moments} using $x^{(1:m)}$.
\State Compute approximations $\hat{M}_v$ and $\hat{\pi}_v$ to $M_v$ and $\pi$ using tensor decomposition.
\State Compute $\sigma$ maximizing $\sum_{j=1}^k \hat{\pi}_{\sigma(j)} \sum_{v=1}^3 (\hat{M}_v)_{j,\sigma(j)}$ using maximum bipartite matching.
\State \textbf{Output}: estimated $\frac{1}{m}\sum_{i=1}^m A(\theta; x^{(i)}) - \sum_{j=1}^k \hat{\pi}_{\sigma(j)} \sum_{v=1}^3 (\hat{M}_v)_{j,\sigma(j)}$.
\end{algorithmic}                                                                  
\end{algorithm}

\section{Extensions}
\label{sec:extensions}

Theorem~\ref{thm:tensor} provides a basic building block 
which admits several extensions to more complex model structures. 
We go over several cases below, omitting most proofs to avoid tedium.
\vskip 0.07in
\begin{extension}[Hidden Markov Model]
\label{ex:hmm}
Most importantly, the latent variable $y$ need not belong to 
a small discrete set; we can handle structured output spaces 
such as a hidden Markov model as long as $p^*$ matches 
the HMM structure. This is a substantial generalization of 
previous work on unsupervised risk estimation, which was 
restricted to multiclass classification.

Suppose that $p_{\theta}(y_{1:T} \mid x_{1:T}) \propto \prod_{t=2}^T f_{\theta}(y_{t-1}, y_t) \cdot \prod_{t=1}^T g_{\theta}(y_t, x_t)$, 
with log-loss $\loss(\theta; x, y) = -\log p_{\theta}(y_{1:T} \mid x_{1:T})$. We can 
exploit the decomposition 
\vskip -0.20in
\begin{equation}
\label{eq:hmm-decomposition}
-\log p_{\theta}(y_{1:T} \mid x_{1:T}) = \sum_{t=2}^T \underbrace{-\log p_{\theta}(y_{t-1}, y_t \mid x_{1:T})}_{\eqdef \ell_t} - \sum_{t=1}^T \underbrace{-\log p_{\theta}(y_t \mid x_{1:T})}_{\eqdef \ell'_t}.
\end{equation}
\vskip -0.10in
Assuming that $p^*$ is Markovian with respect to $y$, each of the 
losses $\ell_t$, $\ell'_t$ satisfies Assumption~\ref{ass:multi-view} 
(see Figure~\ref{fig:hmm}; for $\ell_t$, the views are $x_{1:t-2}, x_{t-1:t}, x_{t+1:T}$ , and for 
$\ell_t'$ they are $x_{1:t-1}$, $x_t$, $x_{t+1:T}$).
We use Theorem~\ref{thm:tensor} to 
estimate each $\bE[\ell_t]$, $\bE[\ell'_t]$ individually, and thus also the full risk $\bE[L]$.
(Note that we actually estimate the risk for $y_{2:T-1} \mid x_{1:T}$ due to 
the $3$-view assumption failing at the boundaries.)

In general, the idea in \eqref{eq:hmm-decomposition} applies to any structured output problem that is a sum of 
local $3$-view structures. It would be interesting to extend our results to 
other structures such as more general graphical models \citepm{chaganty2014graphical} and parse 
trees \citepm{hsu12identifiability}.
\end{extension}
\vskip 0.07in
\begin{extension}[Exponential Loss]
We can also relax the additivity condition $\loss = A - f_1 - f_2 - f_3$.
For instance, suppose 
$\loss(\theta; x,y) = \exp(-\theta^{\top}\sum_{v=1}^3 \phi_v(x_v,y))$ is the exponential loss. 
We can use Theorem~\ref{thm:tensor} to estimate 
the matrices $M_v$ corresponding to $f_v(\theta; x_v, y) = \exp(-\theta^{\top}\phi_v(x_v,y))$. 
Then
\vskip -0.15in
\begin{equation}
\risk(\theta) = \bE\left[\prod_{v=1}^3 f_v(\theta; x_v,y)\right] = \sum_{j} \pi_j \prod_{v=1}^3 \bE\left[f_v(\theta; x_v,j) \mid y=j\right]
\end{equation}
\vskip -0.1in
by conditional independence.
Therefore, the risk can be estimated as $\sum_j \pi_j \prod_{v=1}^3 (M_v)_{j,j}$.
%
More generally, it suffices to have
$\loss(\theta; x,y) = A(\theta; x) + \sum_{i=1}^n \prod_{v=1}^3 f_i^v(\theta; x_v, y)$ 
for some functions $f_i^v$. 
\end{extension}
\vskip 0.05in
\begin{extension}[Mediating Variable]
\label{ex:latent}
Assuming that $x_{1:3}$ are independent conditioned only on $y$ 
may not be realistic; there might be multiple 
subclasses of a label (e.g., multiple ways to write the digit $4$)
which would induce systematic correlations across views. To address 
this, we show that independence 
need only hold conditioned on a mediating variable $z$, rather than on
the label $y$ itself.

Let $z$ be a refinement of $y$ (in the sense that $z \to y$ is deterministic) 
which takes on $k'$ values, and suppose that the views 
$x_1$, $x_2$, $x_3$ are independent conditioned on $z$, as in 
Figure~\ref{fig:latent}.
Then we can estimate the risk as long as we can extend the loss vector $h_v = (f_v)_{i=1}^k$ 
to a function $h'_v : \sX_v \to \bR^{k'}$, such that $h'_v(x_v)_i = f_v(x_v,i)$ and 
the matrix ($M'_v)_{ij} = \bE[h'_v(x_v)_i \mid z=j]$ has full rank.
The reason is that we can recover the matrices $M'_v$, and then, letting $r$ be the 
map from $z$ to $y$, we can express the risk as 
$R(\theta) = \bE[A(\theta; x)] + \sum_{j=1}^{k'} p^*(z=j)\sum_{v=1}^3 (M'_v)_{r(j), j}$.
\end{extension}

%

\paragraph{Summary.} Our framework applies to the log loss and exponential 
loss; to hidden Markov models; and to cases where there are latent variables 
mediating the independence structure.

\section{From Estimation to Learning}
\label{sec:learning}

We now turn our attention to unsupervised learning, i.e., minimizing $R(\theta)$ 
over $\theta \in \bR^d$. Unsupervised learning is impossible without some additional 
information, since even if we could learn the $k$ classes, we wouldn't know which 
class had which label. Thus we assume that we have a small amount of information to break 
this symmetry, in the form of a \emph{seed model} $\theta_0$:
\begin{assumption}[Seed Model]
\label{ass:seed}
We have access to a ``seed model'' $\theta_0$ such that 
$\riski(\theta_0) = \risk(\theta_0)$.
\end{assumption}
Assumption~\ref{ass:seed} merely asks for $\theta_0$ to be 
aligned with the true labels on average.
We can obtain $\theta_0$ from 
a small amount of labeled data (semi-supervised learning) or 
by training in a nearby domain (domain adaptation).
We define $\gap(\theta_0)$ to be the difference between 
$\risk(\theta_0)$ and the next smallest permutation of the classes, 
which will affect the difficulty of learning.

For simplicity we will focus on the case of logistic regression, 
and show how to learn given only Assumptions~\ref{ass:multi-view} 
and \ref{ass:seed}. However, our algorithm extends to general losses, 
as we show in Section~\ref{sec:learning-extensions}.

\paragraph{Learning from moments.}
Note that for logistic regression (Examples~\ref{ex:logistic}), the unobserved components of $\loss(\theta; x,y)$ are linear 
in the sense that $f_v(\theta; x_v, y) = \theta^{\top}\phi_v(x_v, y)$ for some 
$\phi_v$. We therefore have
\vskip -0.16in
\begin{equation}
\label{eq:quasi-linear}
\risk(\theta) = \bE[A(\theta; x)] - \theta^{\top}\bar{\phi}, \text{ where } \bar{\phi} \eqdef \sum_{v=1}^3 \bE[\phi_v(x_v,y)].
\end{equation}
\vskip -0.07in
From \eqref{eq:quasi-linear}, we see that it suffices to 
estimate $\bar{\phi}$, after which all terms on the right-hand-side of 
\eqref{eq:quasi-linear} are known. 
Given an approximation $\hat{\phi}$ to $\bar{\phi}$ (we will show how to obtain $\hat{\phi}$ below), 
we can learn a near-optimal $\theta$ by solving the following 
convex optimization problem:
\begin{equation}
\label{eq:optimization}
\hat{\theta} = \argmin_{\|\theta\|_2 \leq \rho} \bE[A(\theta; x)] - \theta^{\top}\hat{\phi}. 
\end{equation}
In practice we would need to approximate $\bE[A(\theta; x)]$ by samples, 
but we ignore this for simplicity (it only contributes lower-order terms 
to the error).
The $\ell^2$-constraint on $\theta$ imparts robustness to errors in $\bar{\phi}$.
In particular (see Section~\ref{sec:robustness-proof} for a proof):
\begin{lemma}
\label{lem:learning}
Suppose 
$\|\hat{\phi} - \bar{\phi}\|_2 \leq \epsilon$. 
Then the output $\hat{\theta}$ from 
\eqref{eq:optimization} satisfies $\risk(\hat{\theta}) \leq \min_{\|\theta\|_2 \leq \rho} \risk(\theta) + 2\epsilon\rho$.
\end{lemma}
Assuming that the optimal parameter $\theta^*$ has $\ell^2$-norm at most $\rho$, 
Lemma~\ref{lem:learning} guarantees 
that $\risk(\hat{\theta}) \leq \risk(\theta^*) + 2\epsilon \rho$. 
We will see below that computing $\hat{\phi}$ requires
$d\cdot \thePoly/\epsilon^2$ samples.

\paragraph{Computing $\hat{\phi}$.}
Estimating $\bar{\phi}$ can be done in a manner similar to estimating 
$\risk(\theta)$ itself. In addition to the conditional risk matrix 
$M_v \in \bR^{k \times k}$, we compute the \emph{conditional moment matrix}
$G_v \in \bR^{dk \times k}$ defined by 
$(G_v)_{i+kr,j} \eqdef \bE[\phi_v(\theta; x_v, i)_r \mid y=j]$. 
We then have $\bar{\phi}_r = \sum_{j=1}^k \pi_j\sum_{v=1}^3 (G_v)_{j+kr,j}$.

We can solve for $G_1$, $G_2$, and $G_3$ using the same tensor algorithm as in 
Theorem~\ref{thm:tensor}. 
Some care is needed to avoid explicitly forming the 
$(kd) \times (kd) \times (kd)$ tensor that would result from the third 
term in \eqref{eq:moments}, as this would require $\oo\p{k^3d^3}$ memory and is 
thus intractable for even moderate values of $d$. We take a standard approach 
based on random projections \citepm{halko2011structure} and described in 
Section 6.1.2 of 
\citetm{anandkumar13tensor}. 
We refer the reader to the aforementioned references for details, 
and cite only the final sample complexity and runtime (see 
Section~\ref{sec:gradient-proof} for a proof sketch):
\begin{theorem}
\label{thm:gradient}
Suppose that Assumption~\ref{ass:multi-view} holds and that 
$\theta_0 \in \Theta_0$. Let $\delta < 1$ and 
$\epsilon < \min(1, \gap(\theta_0))$. 
Then, given $m = \thePoly \cdot \frac{\log(2/\delta)}{\epsilon^2}$
samples, where $\lambda$ and $\tau$ are as defined in 
\eqref{eq:estimation-complexity}, 
with probability $1-\delta$ we can recover $M_v$ and $\pi$ to error 
$\epsilon$, and $G_v$ to error $(B/\tau)\epsilon$, 
where $B^2 = \bE[\sum_{i,v} \|\phi_v(x_v,i)\|_2^2]$
measures the $\ell^2$-norm of the features. The 
algorithm runs in time $\oo\p{d \p{m + \poly(k)}}$, and the errors are in
Frobenius norm for $M$ and $G$, and $\infty$-norm for $\pi$.
\end{theorem}
\paragraph{Interpretation.} Whereas before we estimated $M_v$ to error 
$\epsilon$, now we estimate $G_v$ (and hence $\bar{\phi}$) 
to error $(B/\tau)\epsilon$. To achieve error 
$\epsilon$ in estimating $G_v$ requires 
$(B/\tau)^2 \cdot \thePoly \frac{\log(2/\delta)}{\epsilon^2}$ samples,
which is $(B/\tau)^2$ times as large as in 
Theorem~\ref{thm:tensor}.
The quantity $(B/\tau)^2$ typically grows as $\oo(d)$, and so the 
sample complexity needed to estimate $\bar{\phi}$ is typically $d$ 
times larger than the sample complexity needed to estimate $\risk$. 
This matches the behavior of the supervised case where we need $d$ times 
as many samples for learning as compared to (supervised) risk estimation of a fixed model.


\paragraph{Summary.}
We have shown how to perform unsupervised logistic 
regression, given only a seed model $\theta_0$.
This enables unsupervised learning under surprisingly weak assumptions 
(only the multi-view and seed model assumptions) even for 
mis-specified models and zero train-test overlap, and 
without assuming covariate shift.
See Section~\ref{sec:learning-extensions} for learning under more general losses.

\section{Experiments}
\label{sec:experiments}

\begin{figure}
\begin{center}
\includegraphics{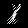}
\includegraphics{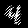}
\includegraphics{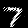}
\hskip 0.3in
\includegraphics{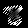}
\includegraphics{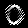}
\includegraphics{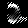}
\end{center}
\caption{A few sample train images (left) and test images (right) from 
the modified MNIST data set.}
\label{fig:mnist}
\vskip -0.08in
\end{figure}

\begin{figure*}
\begin{center}
\begin{subfigure}[t]{0.32\textwidth}
\includegraphics[width=0.95\textwidth]{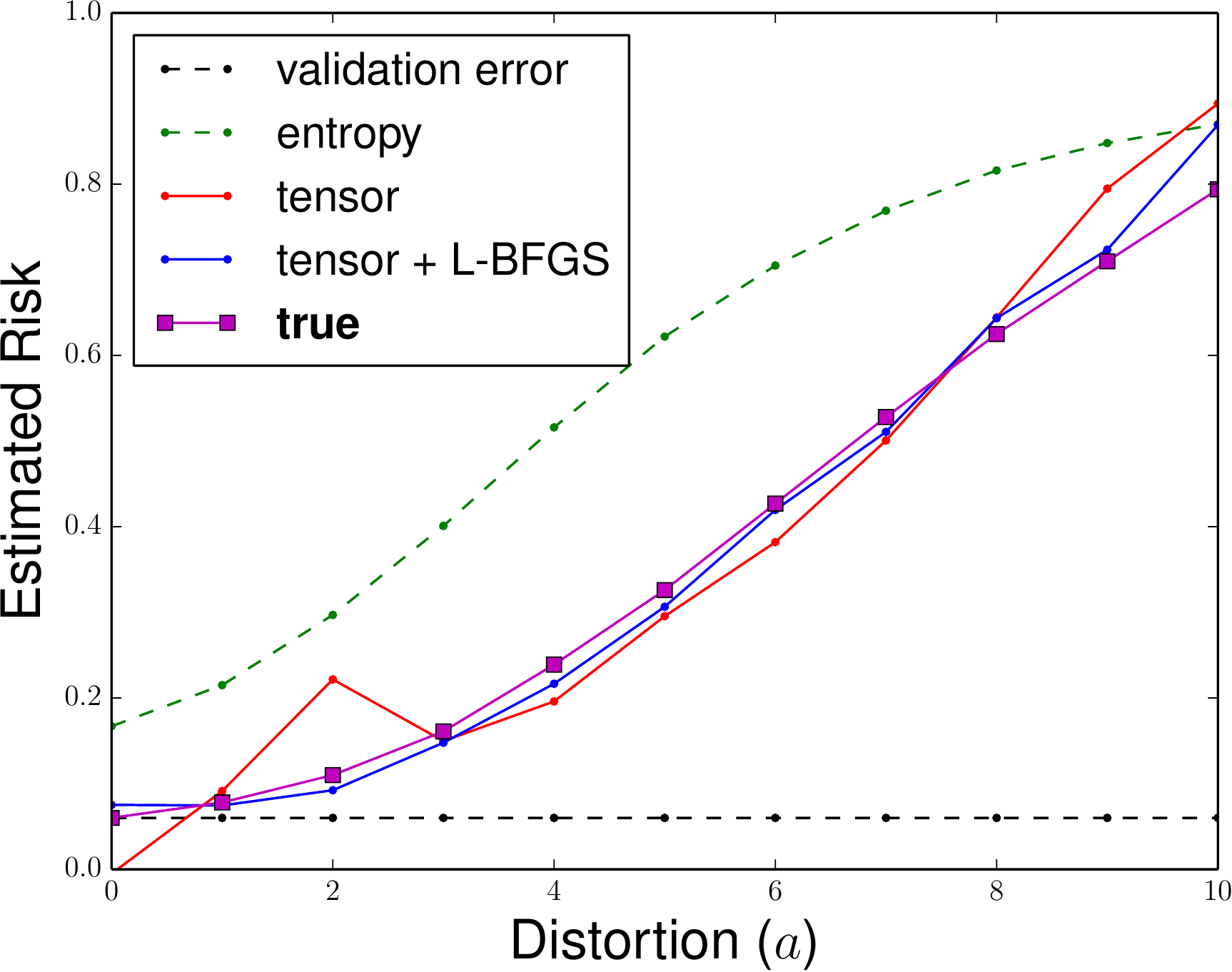}
\vskip -0.03in
\caption{}
\label{fig:results-estimation}
\end{subfigure}
\begin{subfigure}[t]{0.32\textwidth}
\includegraphics[width=0.95\textwidth]{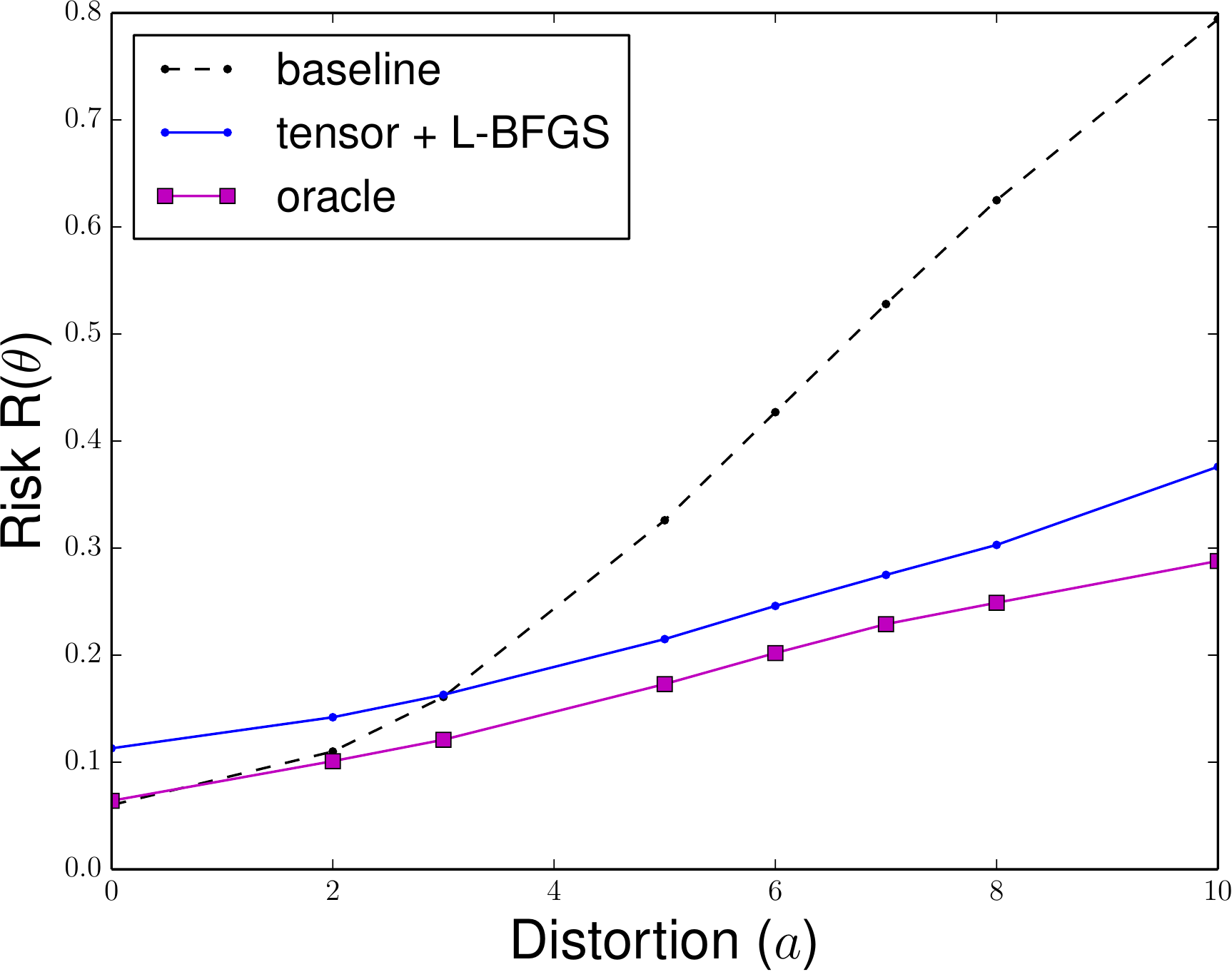}
\vskip -0.03in
\caption{}
\label{fig:results-transfer}
\end{subfigure}
\begin{subfigure}[t]{0.32\textwidth}
\includegraphics[width=0.95\textwidth]{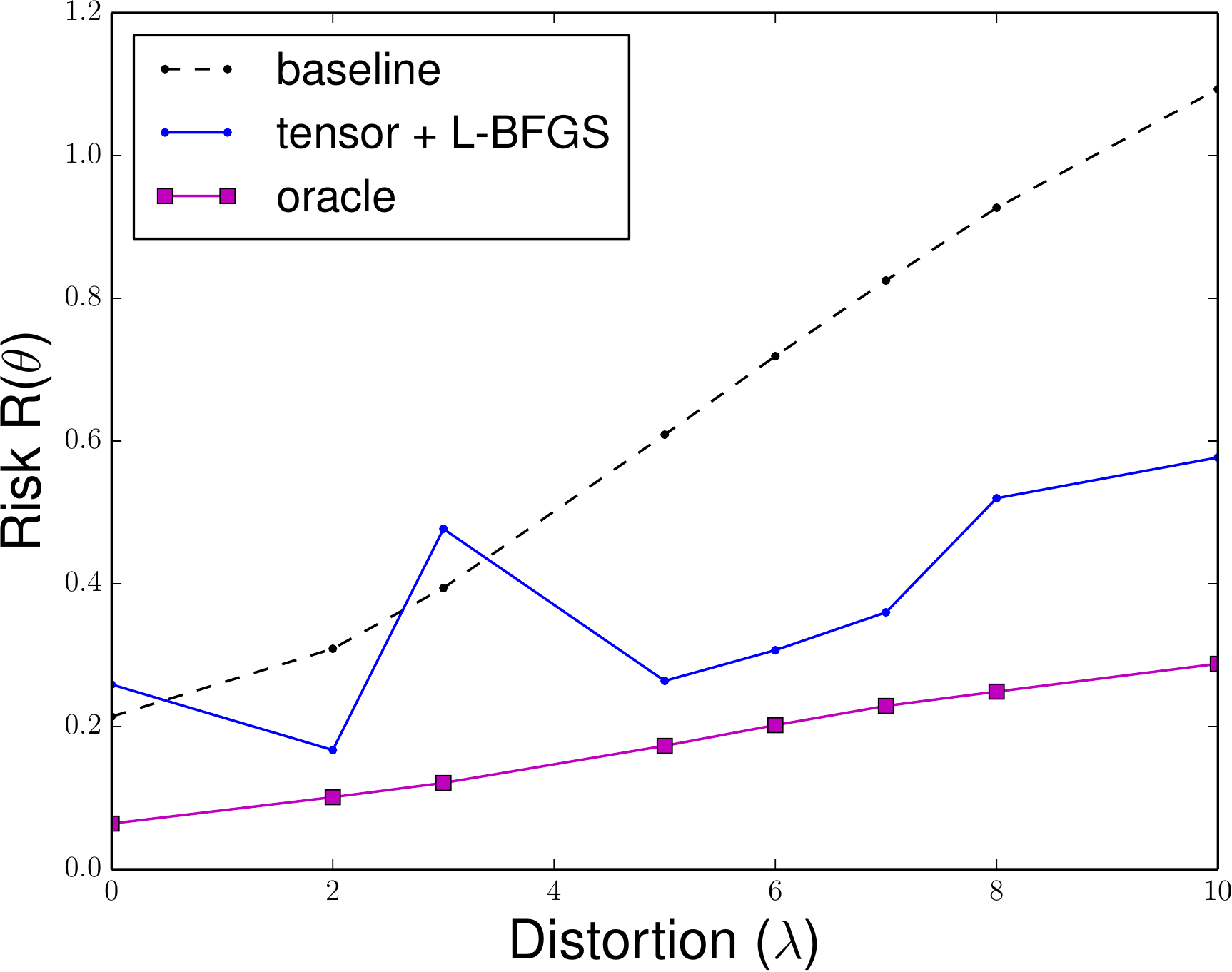}
\vskip -0.03in
\caption{}
\label{fig:results-semisupervised}
\end{subfigure}
\end{center}
\vskip -0.15in
\caption{
  Results on the modified MNIST data set. (a) 
Risk estimation 
for varying degrees of distortion $a$. (b) Domain adaptation
with $10,000$ training and $10,000$ test examples. (c)
Domain adaptation with $300$ training and $10,000$ test examples.}
\end{figure*}

To better understand the behavior of our algorithms, we perform experiments 
on a version of the MNIST data set that is 
modified to ensure that the $3$-view assumption holds. 
To create an image, we sample a class in $\{0,\ldots,9\}$, then 
sample $3$ images at random from that class, letting every 
third pixel come from the respective image. This guarantees 
that there will be $3$ conditionally independent views.
To explore train-test variation, we dim pixel $p$ in the image by 
$\exp\p{a\p{\|p-p_0\|_2 - 0.4}}$, where $p_0$ is the 
image center and the distance is normalized to have maximum 
value $1$. We show example images for $a = 0$ (train) and 
$a = 5$ (a possible test distribution) in Figure~\ref{fig:mnist}.

\textbf{Risk estimation.}
We use unsupervised risk estimation (Theorem~\ref{thm:tensor}) 
to estimate the risk of a model trained on $a = 0$ and tested 
on various values of $a \in [0,10]$. 
We trained the model with AdaGrad \citepa{duchi10adagrad} 
on $10,000$ training examples, and used 
$10,000$ test examples to estimate the risk.
To solve for $\pi$ and $M$ in \eqref{eq:moments}, we first 
use the tensor power method implemented by \citetm{chaganty13regression} to initialize, 
and then locally minimize a weighted 
$\ell^2$-norm of the moment errors with L-BFGS.
For comparison, we compute the validation error for $a = 0$ (i.e., assume 
train = test), as well as the predictive entropy 
$\sum_j -p_{\theta}(j \mid x)\log p_{\theta}(j \mid x)$ on the test set 
(i.e., assume the predictions are well-calibrated).
The results are shown in Figure~\ref{fig:results-estimation}; both the 
tensor method in isolation and tensor + L-BFGS estimate the risk 
accurately, with the latter performing slightly better.

\textbf{Domain adaptation.}
We next evaluate our learning algorithm. For $\theta_0$ we used 
the trained model at $a = 0$, and constrained $\|\theta\|_2 \leq 10$ in \eqref{eq:optimization}.
The results are shown in Figure~\ref{fig:results-transfer}. 
For small values of $a$, 
our algorithm performs worse than 
the baseline of directly using $\theta_0$.
However, our algorithm is far more robust 
as $a$ increases, and tracks the performance of an oracle 
that was trained on the same distribution as the test examples.

\textbf{Semi-supervised learning.}
Because we only need to provide our algorithm with a seed model, we 
can perform \emph{semi-supervised domain adaptation} --- train a model 
from a small amount of labeled data at $a = 0$, then use unlabeled 
data to learn a better model on a new distribution.
Concretely, we obtained $\theta_0$ from only $300$ labeled examples.
Tensor decomposition sometimes led 
to bad initializations in this regime, in which case we obtained a 
different $\theta_0$ by 
training with a smaller step size. The results are shown in 
Figure~\ref{fig:results-semisupervised}. Our algorithm generally 
performs well, but has higher variability than before, seemingly 
due to higher condition number of the matrices $M_{v}$.

\textbf{Summary.} Our experiments show that given $3$ views,
we can estimate the risk and perform 
domain adaptation, even from a small amount of labeled data. 

\section{Discussion}
\label{sec:discussion}

We have presented a method for estimating the 
risk from unlabeled data, which relies only on
conditional independence structure and hence makes 
no parametric assumptions about the true distribution. 
Our approach applies to a large family of losses and extends 
beyond classification tasks to hidden Markov models. We can 
also perform unsupervised learning given only a seed model that 
can distinguish between classes in expectation; 
the seed model can be trained on a related domain, 
on a small amount of labeled data, or any combination of 
the two, and thus provides a pleasingly general formulation
highlighting the similarities between domain adaptation and 
semi-supervised learning.

Previous approaches to domain adaptation and 
semi-supervised learning have also exploited multi-view structure. 
Given two views, \citetm{blitzer2011domain} perform
domain adaptation with zero source/target overlap (covariate shift 
is still assumed). Two-view approaches (e.g. co-training and CCA) are also 
used in semi-supervised learning 
\citepa{blum98cotraining,ando2007two,kakade2007multi,balcan2010discriminative}. 
These methods all assume some form of low noise or low regret, 
as do, e.g., transductive SVMs \citepa{joachims1999transductive}.
By focusing on the central problem of risk estimation, 
our work connects multi-view learning approaches
for domain adaptation and semi-supervised learning,
and removes covariate shift and low-noise/low-regret assumptions 
(though we make stronger independence assumptions, and specialize to 
discrete prediction tasks).


In addition to reliability and unsupervised learning, 
our work is motivated by the desire to build 
\emph{machine learning system with contracts}, a 
challenge recently posed by \citetm{bottou2015two};
the goal is for machine learning systems to
satisfy a well-defined input-output contract in analogy 
with software systems \citepm{sculley2015hidden}. 
Theorem~\ref{thm:tensor} provides the contract 
that under the $3$-view assumption the test error 
is close to our estimate of the test error;
this contrasts with the typical weak contract that if train and 
test are similar, then the test error is close to the training error.
One other interesting contract is given by \citetm{shafer2008tutorial}, 
who provide prediction \emph{regions} that 
contain the true prediction with probability $1-\epsilon$ in the online 
setting, even in the presence of model mis-specification.

The most restrictive part of our framework is the three-view 
assumption, which is inappropriate if the views are not 
completely independent or if the data have structure 
that is not captured in terms of multiple views. Since 
\citeta{balasubramanian2011unsupervised} obtain results under Gaussianity 
(which would be implied by many somewhat dependent views),
we are optimistic that unsupervised risk estimation is possible for a wider 
family of structures. Along these lines, we end with the following 
questions:

\textbf{Open question.}
In the $3$-view setting, suppose the views 
are not completely independent. Is it still possible 
to estimate the risk? 
How does the degree of dependence 
affect the number of views needed?

\textbf{Open question.}
Given only two independent views, can one obtain an 
upper bound on the risk $\risk(\theta)$?

The results of this paper have caused us to adopt 
the following perspective:~To handle unlabeled data, we should 
make \emph{generative} structural assumptions, but still optimize 
\emph{discriminative} model performance. This hybrid approach allows 
us to satisfy the traditional machine learning goal of predictive
accuracy, while handling lack of supervision and under-specification 
in a principled way. Perhaps, then, what is needed for learning is 
to understand the \emph{structure} of a domain.

{\small
\bibliographystyle{plainnat}
\bibliography{refdb/all}
}

\newpage
\onecolumn
\appendix
\section{Details of Computing $\riski$ from $M$ and $\pi$}
\label{sec:matching-details}

In this section we show how, given $M$, and $\pi$, we can 
efficiently compute
\[ \riski(\theta) = \bE[A(\theta; x)] - \max_{\sigma \in \Sym(k)} \sum_{j=1}^k \pi_{\sigma(j)} \sum_{v=1}^3 (M_v)_{j,\sigma(j)}. \]
The only bottleneck is the maximum over $\sigma \in \Sym(k)$, which would 
na\"{i}vely require considering $k!$ possibilities. However, we can 
instead cast this as a form of maximum matching. In particular, form 
the $k \times k$ matrix
\[ X_{i,j} = \pi_i\sum_{v=1}^3 (M_v)_{j,i}. \]
Then we are looking for the permutation $\sigma$ such that 
$\sum_{j=1}^k X_{\sigma(j),j}$ is maximized. If we consider 
each $X_{i,j}$ to be the weight of edge $(i,j)$ in a complete 
bipartite graph, then this is equivalent to asking for a matching 
of $i$ to $j$ with maximum weight, hence we can maximize over $\sigma$ 
using any maximum-weight matching algorithm such as the Hungarian algorithm, 
which runs in $\oo\p{k^3}$ time 
\citepa{tomizawa1971techniques,edmonds1972theoretical}. 

\section{Proof of Theorem~\ref{thm:tensor}}
\label{sec:tensor-proof}

\paragraph{Preliminary reductions.}
Our goal is to estimate $M$ and $\pi$ to error $\epsilon$ (with probability 
of failure $1-\delta$) in $\thePoly \cdot \frac{\log(1/\delta)}{\epsilon^2}$ samples.
Note that if we can estimate $M$ and $\pi$ to error 
$\epsilon$ with any fixed probability of success $1-\delta_0 \geq \frac{3}{4}$, then 
we can amplify the probability of success to $1-\delta$ at the cost of 
$\oo\p{\log(1/\delta)}$ times as many samples (the idea is to make several independent 
estimates, then throw out any estimate that is more than $2\epsilon$ away from at least 
half of the others; all the remaining estimates will then be within distance $3\epsilon$ 
of the truth with high probability).

\paragraph{Estimating $M$.} 
Estimating $\pi$ and $M$ is mostly an exercise in interpreting 
Theorem 7 of \citeta{anandkumar12moments}, which we recall below, 
modifying the statement slightly to fit our language. Here 
$\kappa$ denotes condition number (which is the ratio of $\sigma_1(M)$ to 
$\sigma_k(M)$, since all matrices in question have $k$ columns).

\begin{theorem}[\citeta{anandkumar12moments}]
\label{thm:anandkumar12}
Let $P_{v,v'} \eqdef \bE[h_{v}(x) \otimes h_{v'}(x)]$, 
and $P_{1,2,3} \eqdef \bE[h_1(x) \otimes h_2(x) \otimes h_3(x)]$. 
Also let $\hat{P}_{v,v'}$ and $\hat{P}_{1,2,3}$ be sample estimates 
of $P_{v,v'}$, $P_{1,2,3}$ that are (for technical convenience) 
estimated from independent samples of size $m$. 
Let $\|T\|_F$ denote the $\ell^2$-norm of $T$ after unrolling $T$ to a vector. 
Suppose that:
\begin{itemize}
\item $\bP\left[\|\hat{P}_{v,v'}-P_{v,v'}\|_2 \leq C_{v,v'} \sqrt{\frac{\log(1/\delta)}{m}} \right] \geq 1-\delta$ for $\{v,v'\} \in \{\{1,2\},\{1,3\}\}$, and 
\item $\bP\left[\|\hat{P}_{1,2,3}-P_{1,2,3}\|_F \leq C_{1,2,3} \sqrt{\frac{\log(1/\delta)}{m}}\right] \geq 1-\delta$.
\end{itemize}
Then, there exists constants $C$, $m_0$, $\delta_0$ such that the following holds:
if $m \geq m_0$ and $\delta \leq \delta_0$ and 
\begin{align*}
\sqrt{\frac{\log(k/\delta)}{m}}& \leq C \cdot \frac{\min_{j \neq j'} \|(M_3^{\top})_j - (M_3^{\top})_{j'}\|_2 \cdot \sigma_k(P_{1,2})}{C_{1,2,3} \cdot k^5 \cdot \kappa(M_1)^4} \cdot \frac{\delta}{\log(k/\delta)} \cdot \epsilon, \\
\sqrt{\frac{\log(1/\delta)}{m}}& \leq C \cdot \min\left\{\frac{\min_{j \neq j'} \|(M_3^{\top})_j - (M_3^{\top})_{j'}\|_2 \cdot \sigma_k(P_{1,2})^2}{C_{1,2} \cdot \|P_{1,2,3}\|_F \cdot k^5 \cdot \kappa(M_1)^4} \cdot \frac{\delta}{\log(k/\delta)}, \frac{\sigma_{k}(P_{1,3})}{C_{1,3}}\right\} \cdot \epsilon,
\end{align*}
then with probability at least $1-5\delta$, we can output 
$\hat{M}_3$ with the 
following guarantee: there exists a permutation $\sigma \in \Sym(k)$ such that
for all $j \in \{1,\ldots,k\}$, 
\[ \|(M_3^{\top})_j - (\hat{M}_3^{\top})_{\sigma(j)}\|_2 \leq \max_{j'} \|(M_3^{\top})_{j'}\|_2 \cdot \epsilon. \]
\end{theorem}
By symmetry, we can use Theorem~\ref{thm:anandkumar12} to recover 
each of the matrices $M_v$, $v=1,2,3$, up to permutation of the columns. 
Furthermore, \citeta{anandkumar12moments} show in Appendix B.4 of their paper 
how to match up the columns of the different $M_v$, so that only a single 
unknown permutation is applied to each of the $M_v$ simultaneously.
We will set $\delta = 1/180$, which yields an overall 
probability of success of $11/12$ for this part of the proof.

We now analyze the rate of convergence implied by Theorem~\ref{thm:anandkumar12}. 
Note that we can take $C_{1,2,3} = \oo\p{\sqrt{\bE[\|h_1\|_2^2\|h_2\|_2^2\|h_3\|_2^2]}}$, 
and similarly $C_{v,v'} = \oo\p{\sqrt{\bE[\|h_v\|_2^2\|h_{v'}\|_2^2]}}$. Then, 
since we only care about polynomial factors, it is enough to note that we 
can estimate the $M_v$ to error $\epsilon$ given $Z/\epsilon^2$ samples, 
where $Z$ is polynomial in the following quantities:
\begin{enumerate}
\item $k$, 
\item $\max_{v=1}^3 \kappa(M_v)$, where $\kappa$ denotes condition number,
\item $\frac{\sqrt{\bE[\|h_1\|_2^2\|h_2\|_2^2\|h_3\|_2^2]}}{\p{\min_{j,j'} \|(M_v^{\top})_j - (M_v^{\top})_{j'}\|_2} \cdot \sigma_{k}(P_{v',v''})}$, where $(v,v',v'')$ is a permutation of $(1,2,3)$, 
\item $\frac{\|P_{1,2,3}\|_2}{\p{\min_{j,j'} \|(M_v^{\top})_j - (M_v^{\top})_{j'}\|_2} \cdot \sigma_{k}(P_{v',v''})}$, where $(v,v',v'')$ is as before, and
\item $\frac{\sqrt{\bE[\|h_v\|_2^2\|h_{v'}\|_2^2]}}{\sigma_{k}\p{P_{v,v'}}}$.
\item $\max_{j,v} \|(M_v^{\top})_j\|_2$.
\end{enumerate}
It suffices to show that each of these quantities are polynomial in 
$k$, $\pi_{\min}^{-1}$, $\tau$, and $\lambda^{-1}$. 

(1) $k$ is trivially polynomial in itself.

(2) Note that $\kappa(M_v) \leq \sigma_{1}(M_v) / \lambda \leq \|M_v\|_F / \lambda$. 
    Furthermore, $\|M_v\|_F^2 = \sum_{j} \|\bE[h_v \mid j]\|_2^2 \leq \sum_j \bE[\|h_v\|_2^2 \mid j] \leq k\tau^2$. 
    In all, $\kappa(M_v) \leq \sqrt{k}\tau/\lambda$, which is polynomial in $k$ and $\tau/\lambda$.

(3) We first note that 
$\min_{j \neq j'} \|(M_v^{\top})_j - (M_v^{\top})_{j'}\|_2 = 
\sqrt{2} \min_{j \neq j'} \|M_v^{\top}(e_j-e_{j'})\|_2/\|e_j-e_{j'}\|_2 \geq \sigma_{k}(M_v)$.
    Also, $\sigma_{k}(P_{v',v''}) = \sigma_{k}(M_{v'}\diag(\pi)M_{v''}) \geq \sigma_{k}(M_{v'}) \pi_{\min} \sigma_{k}(M_{v''})$. 
    We can thus upper-bound the quantity in (3.) by 
\begin{align*}
\frac{\sqrt{\bE[\|h_1\|_2^2\|h_2\|_2^2\|h_3\|_2^2]}}{\sqrt{2}\pi_{\min}\sigma_{k}(M_1)\sigma_{k}(M_2)\sigma_{k}(M_3)} 
  &\leq \frac{\tau^3}{\sqrt{2}\pi_{\min}\lambda^3}, 
\end{align*}
which is polynomial in $\pi_{\min}^{-1}$, $\tau/\lambda$.

(4) We can perform the same calculations as in (3), but now we have to bound $\|P_{1,2,3}\|_2$. However, it is easy to see that
\begin{align*}
\|P_{1,2,3}\|_2 &= \sqrt{\|\bE[h_1 \otimes h_2 \otimes h_3]\|_2^2} \\
 &\leq \sqrt{\bE[\|h_1 \otimes h_2 \otimes h_3\|_2^2]} \\
 &= \sqrt{\bE[\|h_1\|_2^2\|h_2\|_2^2\|h_3\|_2^2]} \\
 &= \sqrt{\sum_{j=1}^k \pi_j \prod_{v=1}^3 \bE[\|h_v\|_2^2 \mid y=j]} \\
 &\leq \tau^3,
\end{align*}
which yields the same upper bound as in (3).

(5) We can again perform the same calculations as in (3), where we now 
only have to deal with a subset of the variables, thus obtaining a bound 
of $\frac{\tau^2}{\pi_{\min}\lambda^2}$.

(6) We have $\|(M_v^{\top})_j\|_2 \leq \sqrt{\bE[\|h_{v}\|_2^2 \mid y=j]} \leq \tau$.

In sum, we have shown that with probability $\frac{11}{12}$ we can estimate 
each $M_v$ to column-wise $\ell^2$ error $\epsilon$ using $\thePoly/\epsilon^2$ samples; since 
there are only $k$ columns, we can make the total (Frobenius) error be at most 
$\epsilon$ while still using $\thePoly/\epsilon^2$ samples.
It now remains to estimate $\pi$. 

\paragraph{Estimating $\pi$.}
This part of the argument follows Appendix B.5 of \citeta{anandkumar12moments}.
Noting that $\pi = M_1^{-1}\bE[h_1]$, we can 
estimate $\pi$ as $\hat{\pi} = \hat{M_1}^{-1}\hE[h_1]$, 
where $\hE$ denotes the empirical expectation. 
Hence, we have
\begin{align*}
\|\pi - \hat{\pi}\|_{\infty} &\leq \left\|(\hat{M_1}^{-1}-M_1^{-1})\bE[h_1] + M_1^{-1}(\hE[h_1]-\bE[h_1]) + (\hat{M_1}^{-1}-M_1^{-1})(\hE[h_1]-\bE[h_1])\right\|_{\infty} \\
  &\leq \underbrace{\|\hat{M_1}^{-1}-M_1^{-1}\|_F}_{(i)}\underbrace{\|\bE[h_1]\|_2}_{(ii)} + \underbrace{\|M_1^{-1}\|_F}_{(iii)}\underbrace{\|\hE[h_1]-\bE[h_1]\|_2}_{(iv)} + \underbrace{\|\hat{M_1}^{-1}-M_1^{-1}\|_F}_{(i)}\underbrace{\|\hE[h_1]-\bE[h_1]\|_2}_{(iv)}.
\end{align*}
We will bound each of these factors in turn:
\begin{itemize}
\item[(i)] $\|\hat{M_1}^{-1}-M_1^{-1}\|_F$: let $E_1 = \hat{M_1} - M_1$, 
which by the previous part satisfies $\|E_1\|_F \leq \sqrt{k}\max_{j}\|(\hat{M}_1^{\top})_j - (M_1^{\top})_j\|_2 = \thePoly / \sqrt{m}$. Therefore:
\begin{align*}
\|\hat{M_1}^{-1}-M_1^{-1}\|_F &\leq \|(M_1 + E_1)^{-1} - M_1^{-1}\|_F \\
 &= \|M_1^{-1}(I+E_1M_1^{-1})^{-1} - M_1^{-1}\|_F \\
 &\leq \|M_1^{-1}\|_F\|(I+E_1M_1^{-1})^{-1}-I\|_F \\
 &\leq k\lambda^{-1}\sigma_{1}\p{I+E_1M_1^{-1})^{-1}-I} \\
 &\leq k\lambda^{-1}\frac{\sigma_{1}(E_1M_1^{-1})}{1 - \sigma_{1}(E_1M_1^{-1})} \\
 &\leq k\lambda^{-2}\frac{\|E_1\|_F}{1-\lambda^{-1}\|E_1\|_F} \\
 &\leq \frac{\thePoly}{1-\thePoly/\sqrt{m}} \cdot \frac{1}{\sqrt{m}}.
\end{align*}
We can assume that $m \geq \thePoly$ without loss of generality 
(since otherwise we can trivially obtain the desired bound on 
$\|\pi-\hat{\pi}\|_{\infty}$ by 
simply guessing the uniform distribution), in which case the above quantity 
is $\thePoly \cdot \frac{1}{\sqrt{m}}$.
\item[(ii)] $\|\bE[h_1]\|_2$: we have $\|\bE[h_1]\|_2 \leq \sqrt{\bE[\|h_1\|_2^2]} \leq \tau$.
\item[(iii)] $\|M_1^{-1}\|_F$: since $\|X\|_F \leq \sqrt{k}\sigma_{1}(F)$, we have 
  $\|M_1^{-1}\|_F \leq \sqrt{k}\lambda^{-1}$.
\item[(iv)] $\|\hE[h_1]-\bE[h_1]\|_2$: with any fixed probability 
  (say $11/12$), this term is $\oo\p{\sqrt{\frac{\bE[\|h_1\|_2^2]}{m}}} = \oo\p{\frac{\tau}{\sqrt{m}}}$.
\end{itemize}
In sum, with probability at least $\frac{11}{12}$ 
all of the terms are $\thePoly$, and at least one factor 
in each term has a $\frac{1}{\sqrt{m}}$ decay. Therefore, we have
$\|\pi-\hat{\pi}\|_{\infty} \leq \thePoly \cdot \sqrt{\frac{1}{m}}$.

Since we have shown that we can estimate each of $M$ and $\pi$
individually with probability $\frac{11}{12}$, 
we can estimate them jointly $\frac{5}{6} > \frac{3}{4}$, 
thus completing the proof.

\section{Proof of Lemma~\ref{lem:learning}}
\label{sec:robustness-proof}
Let $B(\rho) = \{\theta \mid \|\theta\|_2 \leq \rho\}$.
First note that $|\theta^{\top}(\hat{\phi} - \bar{\phi})| \leq 
\|\theta\|_2\|\hat{\phi}-\bar{\phi}\|_2 \leq \epsilon\rho$ 
for all $\theta \in B(\rho)$.
Letting $\tilde{\theta}$ denote the minimizer of 
$R(\theta)$ over $B(\rho)$, we obtain
\begin{align}
R(\hat{\theta}) &= \bE[A(\hat{\theta}; x)] -\hat{\theta}^{\top}\bar{\phi} \\ 
 &\leq \bE[A(\hat{\theta}; x)] - \hat{\theta}^{\top}\hat{\phi} + \epsilon\rho \\
 &\leq \bE[A(\tilde{\theta}; x)] - \tilde{\theta}^{\top}\hat{\phi} + \epsilon\rho \\
 &\leq \bE[A(\tilde{\theta}; x)] - \tilde{\theta}^{\top}\bar{\phi} +2\epsilon\rho \\
 &= R(\tilde{\theta}) + 2\epsilon\rho,
\end{align}
as claimed.

\section{Proof of Theorem~\ref{thm:gradient}}
\label{sec:gradient-proof}

We note that Theorem 7 of \citeta{anandkumar12moments} (and hence Theorem~\ref{thm:tensor} above) 
does not require that the $M_v$ be $k \times k$, but only that they have $k$ \emph{columns} (the 
number of rows can be arbitrary). It thus applies for any matrix $M'_v$, where the $j$th columns of 
$M'_v$ is equal to $\bE[h'_v(x_v) \mid j]$ for some $h_v : \sX_v \to \bR^{d'}$.
In our specific case, we will take $h' : \sX_v \to \bR^{k(d+1)}$, where the first 
$k$ coordinates of $h'(x_v)$ are equal to $h(x_v)$ (i.e., $(f_v(x_v,i))_{i=1}^k$), and 
the remaining $kd$ coordinates of $h'(x_v)$ are equal to $\frac{\tau}{B}\frac{\partial}{\partial \theta_r}f_v(\theta; x_v, i)$ 
as in the definition of $G_v$, where the difference is that we have scaled by a factor of 
$\frac{\tau}{B}$. Note that in this case $M'_v = \left[ \begin{array}{c} M_v \\ \frac{\tau}{B}G_v \end{array} \right]$.
We let $\lambda'$ and $\tau'$ denote the values of $\lambda$ and $\tau$ for $M'$ and $h'$.

Since $M_v$ is a submatrix of $M'_v$, we have $\sigma_k(M'_v) \geq \sigma_k(M_v)$, 
so $\lambda' \geq \lambda$.
On the other hand, 
\begin{align}
\tau' &= \bE[\sum_v \|h'_v(x_v)\|_2^2]\\
 &= \bE[\sum_v \|h_v(x_v)\|_2^2 + \frac{\tau^2}{B^2}\sum_{v,i} \|\nabla_{\theta}f_v(\theta; x_v,i)\|_2^2] \\
 &\leq \tau^2 + \frac{\tau^2}{B^2}\bE[\sum_{v,i}\|\nabla_{\theta}f_v(\theta; x_v,i)\|_2^2] \\
 &= 2\tau^2,
\end{align}
so $\tau' \leq \sqrt{2}\tau$.
Since $(\lambda')^{-1} = \oo(\lambda^{-1})$ and $\tau' = \oo(\tau)$, 
we still obtain a sample complexity of 
$\thePoly \cdot \frac{\log(2/\delta)}{\epsilon^2}$.
Since $\theta_0 \in \Theta_0$ by assumption, we can recover the correct permutation 
of the columns of $M_v$ (and hence also of $G_v$, since they are permuted in the same way), 
which completes the proof.

\section{Learning with General Losses}
\label{sec:learning-extensions}

In Section~\ref{sec:learning}, we formed the conditional moment matrix 
$G_v$, which stored the conditional expectation 
$\bE[\phi_v(x_v,i) \mid y=j]$ for each $j$ and $i$. However, there 
was nothing special about computing $\phi$ (as opposed to some other 
moments), and for general losses can form the conditional gradient 
matrix $G_v(\theta)$, defined by
\begin{equation}
\label{eq:G'}
G_v(\theta)_{i+kr,j} = \bE\left[\frac{\partial}{\partial \theta_r}f_v(\theta; x_v, i) \mid y=j\right].
\end{equation}
Theorem~\ref{thm:gradient} applies identically to the matrix $G_v(\theta)$ 
at any fixed $\theta$. We can then compute the gradient $\nabla_{\theta}R(\theta)$ using the relationship
\begin{equation}
\frac{\partial}{\partial \theta_r}R(\theta) = \bE\left[\frac{\partial}{\partial \theta_r}A(\theta; x)\right] - \sum_{j=1}^k \pi_j \sum_{v=1}^3 G_v(\theta)_{j+kr,j}.
\end{equation}
For clarity, we also use $M_v(\theta)$ to denote the conditional risk matrix 
at a value $\theta$. To compute the gradient $\nabla_{\theta}R(\theta)$, we 
jointly estimate $M_v(\theta_0)$ and $G_v(\theta)$ (note the differing arguments 
of $\theta_0$ vs. $\theta$). Since the seed model assumption 
(Assumption~\ref{ass:seed}) allows us to recover the correct column permutation 
for $M_v(\theta_0)$, estimating $G_v(\theta)$ jointly with $M_v(\theta_0)$ 
ensures that we recover the correct column permutation for $G_v(\theta)$ as well.

The final ingredient is any gradient descent procedure that is robust to errors 
in the gradient (so that after $T$ steps with error $\epsilon$ on each step, 
the total error is $\oo(\epsilon)$ and not $\oo(\epsilon T)$). Fortunately, 
this is the case for many gradient descent algorithms, including any 
algorithm that can be expressed as mirror descent (we omit the details because 
they are somewhat beyond our scope, but refer the reader to
Lemma 21 of \citepa{steinhardt2015memory} for a proof of this in the case 
of exponentiated gradent).

The general learning algorithm is given in Algorithm~\ref{alg:learning-general}:

\begin{algorithm}[h!]
\caption{General algorithm for learning via gradient descent.}
\label{alg:learning-general}                                       
\begin{algorithmic}[1]                                        
\State Parameters: step size $\eta$
\State Input: unlabeled samples $x^{(1)},\ldots,x^{(m)} \sim p^*(x)$, seed model $\theta_0$
\State $z^{(1)} \gets 0 \in \bR^d$
\For{$t=1$ {\bfseries to} $T$}
\State $\theta^{(t)} \gets \argmin_{\theta} \frac{1}{2\eta}\|\theta-\theta_0\|_2^2 - \theta^{\top}z$
\State Compute $(M_v^{(t)}, G_v^{(t)})$ by jointly estimating $M_v(\theta_0), G_v(\theta)$ from $x^{(1:m)}$.
\For{$r=1$ {\bfseries to} $d$}
\State $g_r \gets \frac{1}{m}\sum_{i=1}^m \frac{\partial}{\partial \theta_r}A(\theta^{(t)}; x^{(i)}) - \sum_{j=1}^k\pi_j \sum_{v=1}^3 (G_v^{(t)})_{j+kr,j}$
\State $z^{(t+1)}_r \gets z^{(t)}_r - g_r$
\EndFor
\EndFor
\State Output $\frac{1}{T}\p{\theta^{(1)}+\cdots+\theta^{(T)}}$.
\end{algorithmic}                                                                  
\end{algorithm}                



\end{document}